\pdfoutput=1

\documentclass[11pt]{article}

\usepackage[final]{acl}

\usepackage{times}
\usepackage{latexsym}
\usepackage{amssymb}
\usepackage[T1]{fontenc}

\usepackage[utf8]{inputenc}

\usepackage{microtype}

\usepackage{inconsolata}

\usepackage{graphicx}
\usepackage{pifont}

\usepackage{booktabs}
\usepackage{multicol}
\usepackage{multirow}
\usepackage{tcolorbox}
\usepackage{colortbl}
\usepackage{util}
\usepackage{enumitem}

\lstdefinestyle{python}{
    language=Python,
    basicstyle=\fontsize{7}{9.5}\ttfamily,
    keywordstyle=\color{blue},
    commentstyle=\color{gray},
    stringstyle=\color{black},
    showstringspaces=false,
    breaklines=true,
    breakindent=0pt,
    breakatwhitespace=false,
    escapeinside={(*@}{@*)}
}

\lstdefinestyle{plain}{
    basicstyle=\fontsize{7}{9.5}\ttfamily,
    keywordstyle=\color{blue},
    commentstyle=\color{gray},
    stringstyle=\color{green},
    showstringspaces=false,
    breaklines=true,
    breakatwhitespace=false,
    breakindent=0pt,
    escapeinside={(*@}{@*)}
}

\newcommand{\ours}{Table-R1}

\title{Can GRPO Boost Complex Multimodal Table Understanding?}

\author{
    Xiaoqiang Kang$^{1,2}$, 
    Shengen Wu$^3$, 
    Zimu Wang$^{1,2}$, 
    Yilin Liu$^4$, 
    Xiaobo Jin$^1$, \\
    \textbf{Kaizhu Huang}$^5$, 
    \textbf{Wei Wang}$^1$, 
    \textbf{Yutao Yue}$^3$, 
    \textbf{Xiaowei Huang}$^2$, 
    \textbf{Qiufeng Wang}$^{1,}$\thanks{Corresponding author.} \\
    $^1$School of Advanced Technology, Xi'an Jiaotong-Liverpool University \\
    $^2$Department of Computer Science, University of Liverpool \\
    $^3$Information Hub, Hong Kong University of Science and Technology (Guangzhou) \\
    $^4$University of Southern California \quad $^5$Duke Kunshan University \\
    \texttt{Xiaoqiang.Kang23@student.xjtlu.edu.cn, Qiufeng.Wang@xjtlu.edu.cn}
}

\begin{document}
\maketitle
\begin{abstract}
Existing table understanding methods face challenges due to complex table structures and intricate logical reasoning. While supervised fine-tuning (SFT) dominates existing research, reinforcement learning (RL), such as Group Relative Policy Optimization (GRPO), has shown promise but struggled with low initial policy accuracy and coarse rewards in tabular contexts. In this paper, we introduce \ours, a three-stage RL framework that enhances multimodal table understanding through: (1) \textbf{Warm-up} that prompts initial perception and reasoning capabilities, (2) \textbf{Perception Alignment GRPO (PA-GRPO)}, which employs continuous Tree-Edit-Distance Similarity (TEDS) rewards for recognizing table structures and contents, and (3) \textbf{Hint-Completion GRPO (HC-GRPO)}, which utilizes fine-grained rewards of residual steps based on the hint-guided question. Extensive experiments demonstrate that \ours{} can boost the model's table reasoning performance obviously on both held-in and held-out datasets, outperforming SFT and GRPO largely. Notably, Qwen2-VL-7B with \ours{} surpasses larger specific table understanding models (e.g., Table-LLaVA 13B), even achieving comparable performance to the closed-source model GPT-4o on held-in datasets, demonstrating the efficacy of each stage of \ours{} in overcoming initialization bottlenecks and reward sparsity, thereby advancing robust multimodal table understanding.
\end{abstract}
 
\section{Introduction}

\label{sec:intro}

Table understanding is regarded as a cornerstone task in NLP and multimodal research, as structured data in the form of tables is pervasive across diverse domains such as scientific research \cite{pmlr-v235-van-breugel24a, li-etal-2024-multimodal-arxiv}, finance \cite{chen2021finqa, katsis2022ait}, and education \cite{lu2023tabmwp,kangTemplateDrivenLLMParaphrasedFramework2025}. This task presents unique challenges due to the complex table structures and intricate logical reasoning in real-world tables \cite{mathur-etal-2024-knowledge, zhao2024tabpedia}. Effectively interpreting and reasoning over tabular data is critical for enhancing information extraction and automating data analysis.

\begin{figure}[t]
    \centering
    \includegraphics[width=\linewidth]{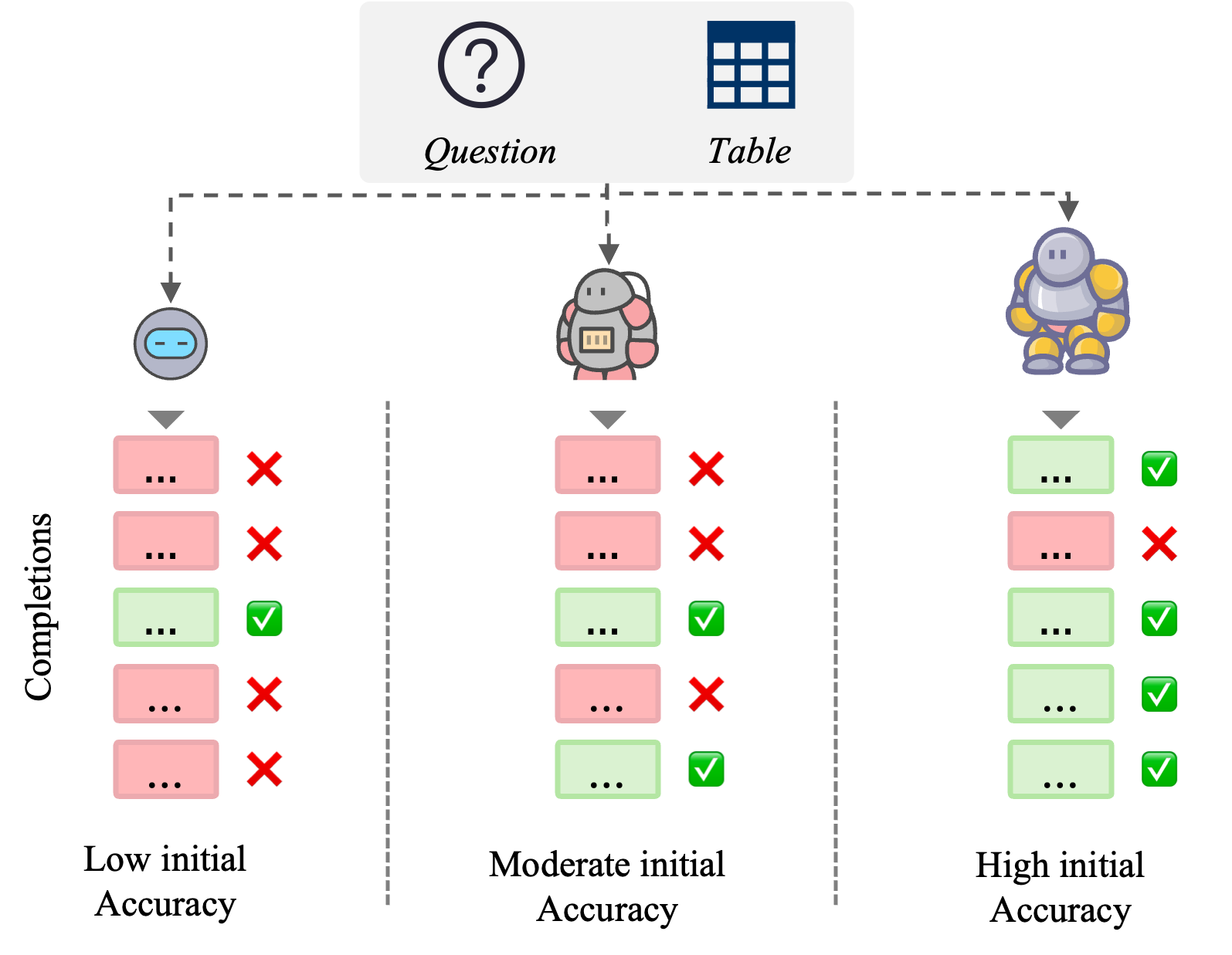}
    \caption{Comparative analysis of different initial policy accuracy in a group. }
    \label{fig:Intro_fig}
    \vspace{-2mm}
\end{figure} 

\begin{table}[t]
\centering
\resizebox{\linewidth}{!}{ 
\begin{tabular}{|c|c|c|c|c|}
\hline
Model & \( A_{\text{init}}\) (\%) & \( A_{\text{final}} \) (\%) & $\Delta A$ & $V_{A_{init}}$ \\ \hline
Qwen2.5-0.5B & 21.4 & 28.7 & 7.3 & 0.168 \\ \hline
Qwen2.5-1.5B & 31.2 & 44.0 & 12.8 & 0.215 \\ \hline
Qwen2.5-3B & 55.2 & 87.6 & \textbf{32.4} & \textbf{0.247} \\ \hline
Qwen2.5-7B & 81.8 & 91.2 & 9.4 & 0.149 \\
\hline
\end{tabular}
}
\caption{Comparative analysis of GRPO performance on TabMWP across Qwen2.5 models of varying scales. $\Delta A = A_{\text{final}} - A_{\text{init}}$ represents the absolute improvement in accuracy, while $\text{V}(A_{\text{init}}) = A_{\text{init}} (1 - A_{\text{init}})$ denotes the variance-based measure of initial policy accuracy. }
\vspace{-4mm}
\label{tab:Intro_table}
\end{table}

Recent research on table understanding has witnessed two predominant paradigms: supervised fine-tuning (SFT) and reinforcement learning (RL). While most work has been largely dominated by SFT \cite{zheng-etal-2024-table-llava,kangTemplateDrivenLLMParaphrasedFramework2025}, these methods suffer from limited generalization when facing unseen table structures or complex reasoning chains \cite{chu2025sftmemorizesrlgeneralizes}. 
In contrast, RL has resurged as a promising paradigm for improving complex reasoning, especially in mathematical tasks. Methods such as Proximal Policy Optimization (PPO, \citealp{PPO}), Direct Preference Optimization (DPO, \citealp{rafailov2023dpo}), and Group Relative Policy Optimization (GRPO, \citealp{shao2024deepseekmath}) demonstrate that RL-based methods can significantly enhance reasoning capabilities. However, the application of RL to multimodal table understanding remains underexplored, despite its potential to address the limitations of SFT-based approaches.  This naturally raises an important research question: \textit{Can RL-based methods such as GRPO be effectively adapted to enhance complex multimodal table understanding for Large Vision Language Models (LVLMs)?}

To accomplish this, we first conduct a preliminary study to investigate the application of GRPO to complex table understanding tasks, identifying a critical dependency on the initial policy's accuracy (see Section \ref{sec:observation}). As shown in Figure \ref{fig:Intro_fig} and Table \ref{tab:Intro_table}, only a policy model with moderate accuracy can produce a balanced mix of correct and incorrect outputs, which is crucial for policy optimization.
This finding highlights a fundamental limitation: the low initial accuracy of the policy model hinders effective back propagation due to the low standard of rewards, ultimately impairing the convergence of the policy model.
Additionally, existing reward functions primarily depend on binary correctness signals. Thus, another challenge is how to devise more fine-grained reward functions tailored for tabular perception and reasoning tasks.
 
\begin{figure*}[t!]
    \centering
    \includegraphics[width=\linewidth]{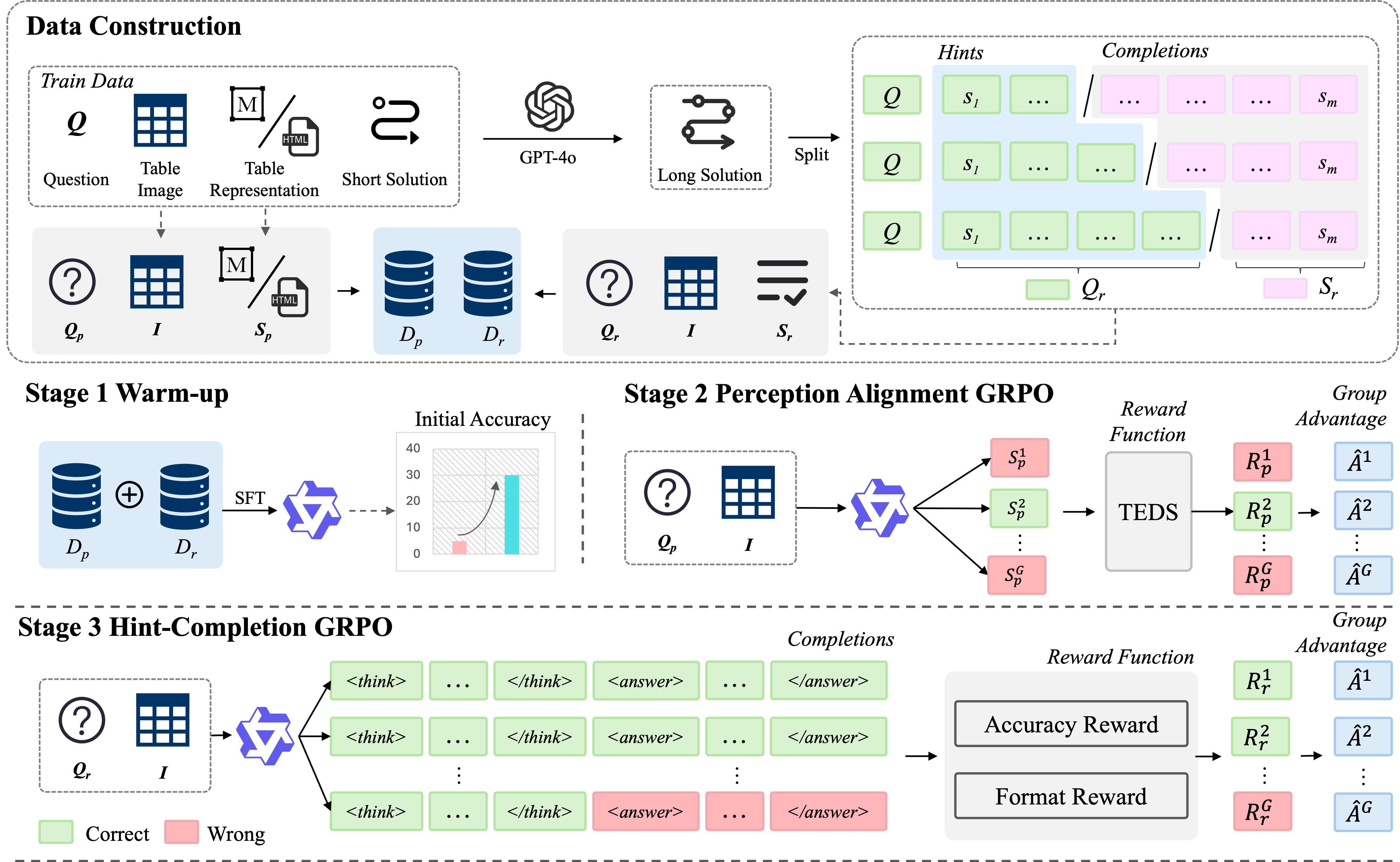}
    \caption{Overall framework of \ours{}. (1) \textbf{Warm-up} establishes foundational capabilities in both visual perception and reasoning. (2) \textbf{PA-GRPO} refines the model's structural understanding by employing TEDS as a continuous reward. (3) \textbf{HC-GRPO} utilizes fine-grained rewards of residual steps based on the hint-guided question.} 
    \label{fig:framework}
    \vspace{-4mm}
\end{figure*}

To address the challenges highlighted, we introduce \ours{}, the first RL-based framework specifically designed for multimodal table understanding. Inspired by the cold-start strategy in DeepSeek-R1 \cite{guo2025deepseekr1}, \ours{} introduces a three-stage framework (see Figure \ref{fig:framework}): (1) \textbf{Warm-up} initializes the model with perception and reasoning capabilities, while also boosting the policy model's initial accuracy. (2) \textbf{Perception Alignment GRPO (PA-GRPO)} employs continuous reward signals, Tree-Edit-Distance-based Similarity (TEDS), for table structure recognition. (3) \textbf{Hint-Completion GRPO (HC-GRPO)} applies reward functions to the residual steps of the hint-guided question, which offers a finer-grained reward than a coarse solution-level reward and further refines the model's reasoning capabilities.

We divide our datasets into two parts: held-in and held-out. The held-in comprises 4 multimodal table understanding tasks for training, whereas similar tasks are set as held-out to assess the model's robustness. 
We validate the effectiveness of each stage of \ours{} and conduct comprehensive experiments compared against baselines. Experimental results indicate that \ours{} consistently outperforms both SFT and GRPO across models of different scales. For Qwen2-VL-7B, \ours{} achieves a 3.93\% improvement over SFT and a 16.38\% improvement over GRPO on held-in, as well as a 7.72\% improvement over SFT and a 8.79\% improvement over GRPO on held-out, significantly surpassing that of models with larger scale (e.g., Table-LLaVA 13B) and matching GPT-4o's performance.

The main contributions of our work are summarized as follows: 
\textbf{(1)} We identify and empirically validate the pivotal limitation of GRPO in table reasoning, that the policy model is sensitive to the initial accuracy; \textbf{(2)} We propose \ours{}, a new three-stage reinforcement learning framework that enables LVLM to improve its perception and reasoning capability for the first time; \textbf{(3)} We conduct comprehensive experiments on six datasets to demonstrate that our framework can obviously surpass both SFT and GRPO, specifically boosting the Qwen2-VL-7B model largely to achieve state-of-the-art performance on several benchmarks. %

\section{Related Work}

\paragraph{Multimodal Table Understanding} is a fundamental task in computer vision and document understanding. Early works have focused on visual table recognition, structure parsing, and content extraction from document images, such as PubTabNet \cite{Xu2020PubTabNet}, FinTabNet \cite{zheng2020FinTabNet}, and TableFormer \cite{yang-etal-2022-tableformer}.

Recent efforts have advanced toward reasoning over visually and contextually rich tables. Representative works include Table-LLaVA \cite{zheng-etal-2024-table-llava}, which augments table inputs with cell-associated images, and TabPedia \cite{zhao2024tabpedia}, which provides a large-scale multimodal table pretraining corpus to improve downstream performance. Multimodal ArXiv \cite{li-etal-2024-multimodal-arxiv} proposes fine-grained reasoning over scientific tables with linked charts and text, while Karma \cite{mathur-etal-2024-knowledge} incorporates symbolic knowledge graphs for better factual alignment. In terms of reasoning supervision, \citet{chengVisionLanguageModelsCan2024} proposes R3V, a self-training framework that iteratively generates and selects chain-of-thought trajectories to improve multimodal question answering on documents and tables.

\paragraph{Reinforcement Learning} (RL), as a machine learning paradigm, aims to learn optimal decision-making by enabling an agent to interact with an environment and relying on reward signals \cite{zhangR1VLLearningReason2025}. In the context of large language models (LLMs), RL is mapped to concrete language generation tasks: the LLM functions as the agent, with user prompts and generated text constituting the environment state, while generating the next token corresponds to the agent’s action \cite{wangReinforcementLearningEnhanced2025}. To facilitate effective training, pre-trained reward models are typically employed. These models automatically evaluate the quality of the generated text based on human preferences or preset criteria, and their outputs serve as rewards that guide the training of the LLM \cite{ouyang2022training}.

In recent years, RL techniques have substantially enhanced the reasoning capabilities of LLMs \cite{luongReFTReasoningReinforced2024, liuVisualRFTVisualReinforcement2025, pengSkyworkR1VPioneering2025}. Numerous studies have adopted appropriate reward functions and policy optimization strategies to reinforce high-quality reasoning paths while penalizing low-quality ones, thereby guiding the models to achieve more coherent and logically structured reasoning trajectories \cite{wangReinforcementLearningEnhanced2025}. For example, \citet{rafailov2023dpo} and \citet{yuanSelfRewardingLanguageModels2025} employ Direct Preference Optimization (DPO), \citet{zhangReSTMCTSLLMSelfTraining2024} utilizes a process reward model to evaluate each reasoning step, and \citet{wangMathShepherdVerifyReinforce2024} leverages both process and outcome reward models simultaneously. Particularly noteworthy is Deepseek-R1, which employs a Group Relative Policy Optimization (GRPO) method to achieve robust reasoning capabilities solely through RL \cite{guo2025deepseekr1}. GRPO replaces the traditional reward function with verifiable rule-based rewards and substitutes multiple sampling for the critic model, directly steering the model to converge on high-quality reasoning strategies without the need for complex reward modeling\cite{shao2024deepseekmath}.

\begin{figure*}[t]
    \centering
    \includegraphics[width=\linewidth]{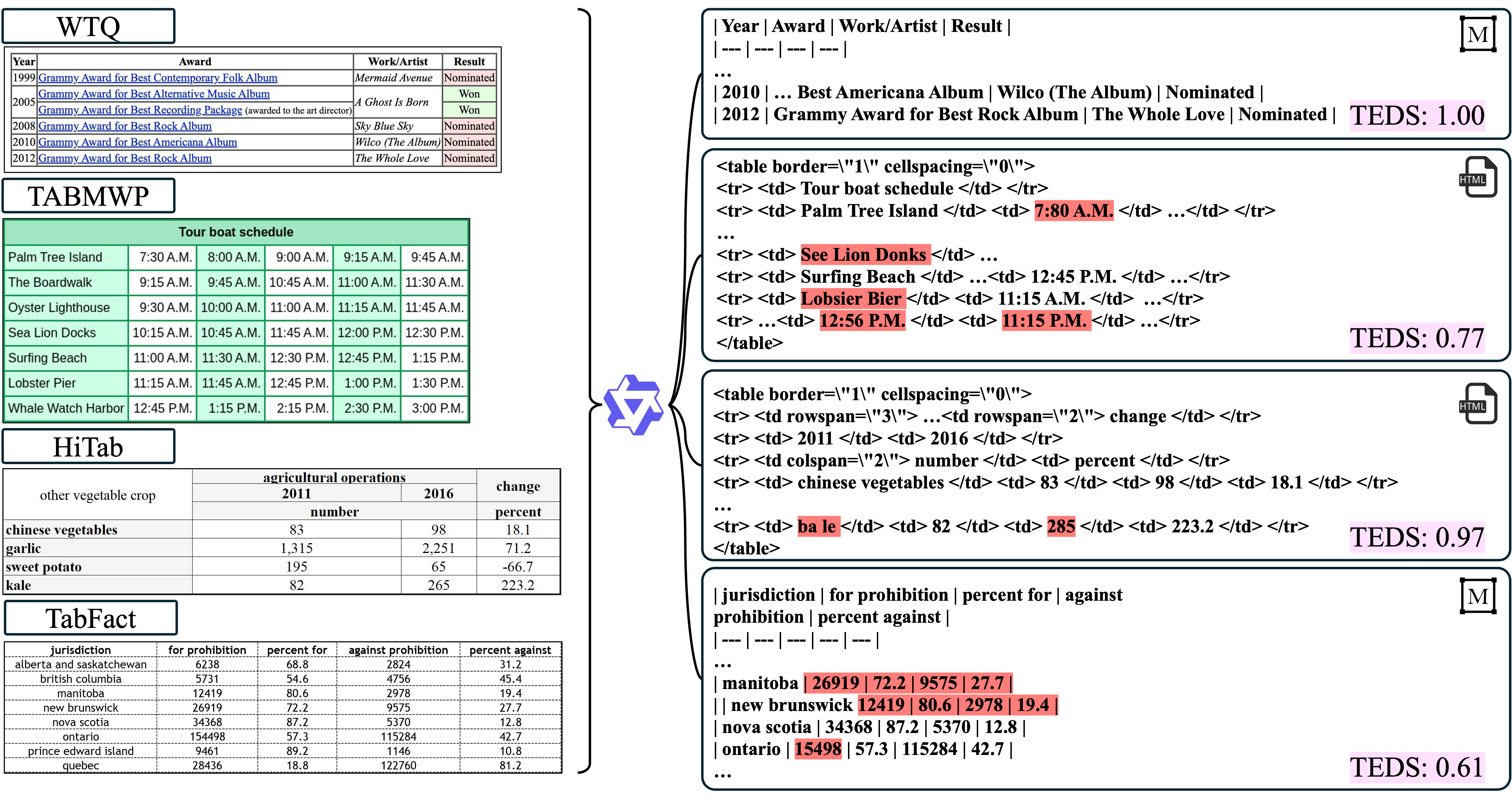}
    \caption{Examples from the datasets used in PA-GRPO. The highlighted red segments indicate the incorrect predictions. TEDS assigns a continuous score to each output, reflecting the similarity to the golden answer.} 
    \vspace{-4mm}
    \label{fig:perception}
\end{figure*}

\section{Observation}
\label{sec:observation}

We first investigate the application of GRPO to complex table understanding tasks, identifying a critical dependency on the initial policy's accuracy. As detailed in Table \ref{tab:Intro_table}, our evaluation on the TabMWP dataset \cite{lu2023tabmwp} reveals a stark performance disparity. Models with either low (e.g., Qwen2.5-0.5B) or high (e.g., Qwen2.5-7B) initial accuracy merely yield gains of 7.3\% and 9.4\%, respectively. In contrast, the Qwen2.5-3B model, starting from a moderate accuracy, achieves a substantial improvement of 32.4\%.
This phenomenon is visually represented in Figure \ref{fig:Intro_fig}, 
where a low-accuracy policy tends to generate mostly incorrect solutions, while a high-accuracy one generates predominantly correct solutions. Only a policy model with moderate accuracy can produce a balanced mix of correct and incorrect outputs, which is crucial for policy optimization.

This behavior can be explained by the variance of the binary reward. Assuming rewards follow a Bernoulli distribution $R \sim \text{Bernoulli}(p)$, where $p = A_{\text{init}}$, the variance is given by $A_{\text{init}}(1 - A_{\text{init}})$. When $A_{\text{init}}$ approaches 0 or 1, the variance approaches zero. This leads to a zero advantage estimate and negligible policy gradients for optimization. Conversely, when $A_{\text{init}} \approx 0.5$ (e.g., Qwen2.5-3B at 55.2\%), variance approaches the theoretical maximum of 0.25, maximizing advantage and providing strong gradients for effective policy optimization.
This finding highlights a fundamental limitation: the low initial accuracy of the policy model hinders effective back propagation due to the low standard of rewards, ultimately impairing the convergence of the policy model. This observation aligns with recent findings emphasizing GRPO's sensitivity to policy initialization \cite{yuDAPOOpenSourceLLM2025, liuUnderstandingR1ZeroLikeTraining2025}.

\section{Methodology}
\subsection{Problem Formulation}
\label{subsec:problem_formulation}
The multimodal tabular input, $(I, Q)$, consists of a question $Q$ and a corresponding table image $I$. The policy model $\pi_{\theta}$ generates a series of actions to generate token sequences, which comprises a step-by-step reasoning trajectory and the final answer. Each action corresponds to generating the next token in the output sequence. For each rollout, the model produces candidate outputs $\{S^1, \dots, S^G\}$ with corresponding rewards $\{R^1, \dots, R^G\}$. The objective is to optimize $\pi_{\theta}$ to maximize the expected cumulative reward $\mathbb{E}_{S\sim\pi_{\theta}(I,Q)}[R(S)]$ by selecting actions that generate high-quality reasoning trajectories.

\subsection{\ours{} Framework}
\label{subsec:table_r1}

As illustrated in Figure \ref{fig:framework}, we initially propose \ours{}, a three-stage training framework for tabular perception and reasoning tasks. (1) \textbf{Warm-up}: Supervised fine-tuning to initialize the model with strong perception and reasoning capabilities. (2) \textbf{Perception Alignment GRPO}: Improves table structure recognition using continuous rewards. (3) \textbf{Hint-Completion GRPO}: Enhances step-by-step reasoning through hint-based completions. The overall training algorithm is shown in Appendix \ref{apd:Pseudocode}.

\paragraph{Warm-up.} As shown in Figure \ref{fig:Intro_fig}, the initial accuracy of the policy model plays a crucial role during the training of GRPO. To address this, we introduce a warm-up stage that significantly boosts the model’s initial perception and reasoning accuracy via SFT. This stage equips the policy model with the ability to both convert images to structured table representations and to generate valid step-wise reasoning paths. During the warm-up stage, the policy model undergoes SFT using the perception task dataset $D_p$ and reasoning task dataset $D_r$, which will be detailed in the following two stages.
The loss function we used here is:
\begin{equation}
\small
\mathcal{L}_{\text{warm-up}}=-\mathbb{E}_{(I,Q,S) \sim D_{p} \cup D_{r}}\left[\sum_{t=1}^{T} \log \pi_{\theta}(s_t|s_{<t})\right].
\label{eq:warmup}
\end{equation}

\paragraph{Perception-Alignment GRPO (PA-GRPO).} In this stage, the model focuses on its ability to recognize patterns and structures. The model extracts structured tabular representations from input images $I$, generating outputs  $S_p$ in either Markdown or HTML format. To enhance the linguistic diversity of instruction, we have constructed 20 distinct instruction variants $Q_p$ for this task, as shown in Figure \ref{fig:question-perception-tr}  in Appendix \ref{sec:Instruction-variants-for-PA-GRPO}. 
The complete dataset, denoted as $D_p$, is a collection of tuples $(I, Q_p, S_p)$, where each tuple consists of a table image $I$, an instruction variant $Q_p$, and the target structured representation $S_p$. Tree-Edit-Distance-based Similarity (TEDS) \cite{zhong2019image} is utilized as a reward. This similarity is calculated based on the tree structure of the table sequence. It assesses both structural similarity and content similarity of the cells between the predicted table $S_p$ and the golden answer $GA$. TEDS is normalized on a scale from 0 to 1, where a score of 1 indicates a perfect match. Several detailed examples are reported in Figure \ref{fig:perception}. Formally, the reward is defined as follows:
\begin{align}
    R_{p} = \text{TEDS}(S_p, GA).
\end{align}
Since this perception task doesn't require a reasoning process, LVLM is expected to provide direct answers, and the complete prompt is displayed in Table \ref{tab:prompt-for-grpo} in Appendix \ref{sec:prompt-for-grpo}.

\paragraph{Hint-Completion GRPO (HC-GRPO).} During this stage, given a question $Q$, the model enhances its reasoning capability by progressively completing the remaining steps to reach the final answer. The residual-step rewards can be more fine-grained than solution-level. Some initial solutions are too brief to be effectively split into two parts, so we employ GPT-4o~\cite{openai2024gpt4technicalreport} to expand short solutions into long reasoning chains $[s_1, s_2, \ldots, s_n]$, where each $s_i$ represents the $i$-th step of the extended solution. The detailed prompts are displayed in Figure \ref{fig:prompt-data-rewriting} in Appendix \ref{sec:prompt-for-data-construction}. 

\begin{table*}[t] %
\renewcommand{\arraystretch}{1.3}
\setlength\tabcolsep{2pt}
\centering
\scalebox{0.75}{
\begin{tabular}{c|c|cccc|cc|cc|c} 
\toprule
\multirow{2}{*}{\textbf{Task Category}} & \multirow{2}{*}{\textbf{Task Name}} & \multirow{2}{*}{\textbf{Dataset}} & \multirow{2}{*}{\textbf{Table Style}} & \multirow{2}{*}{\textbf{Source}} & \multirow{2}{*}{\textbf{Held-in}} & \multicolumn{2}{c|}{\textbf{ Original}} & \multicolumn{2}{c|}{\textbf{ Sampled}} & \multirow{2}{*}{\textbf{Avg. Pixel}}  \\  %
\cmidrule{7-10}
 &  &  &  &  &  & \# T & \# Q & \# T & \# Q & \\ 
\hline
\multirow{6}{*}{\begin{tabular}[c]{@{}c@{}}Table\\Question \\Answering\\(TQA)\end{tabular}} & Flat TQA & WTQ (\citeyear{WTQ}) & W & Wikipedia & Yes & 1.6K & 17K & 1.6K & 8K  & 1992$\times$1116 \\ 
\cmidrule{2-11}
 & \multirow{1}{*}{Hierarchical TQA} & HiTab (\citeyear{HiTab}) & E & \begin{tabular}[c]{@{}c@{}}Wikipedia ~\\Goverment Reports\end{tabular} & Yes & 3K & 8K & 3K & 8K & 3057$\times$793 \\ 
\cmidrule{2-11}
 & \multirow{1}{*}{\begin{tabular}[c]{@{}c@{}}Tabular\\Numerical Reasoning\end{tabular}} & TabMWP (\citeyear{lu2023tabmwp}) & W & Math Exams & Yes & 30K & 30K & 8K & 8K & 267$\times$191 \\ 
 &  & TAT-QA (\citeyear{zhu2021tatqa}) & M & Financial Reports & No & 1.7K & 5.9K & / & / & 2446$\times$1141 \\  
\midrule
\multirow{1}{*}{\begin{tabular}[c]{@{}c@{}}Table Fact \\Verification (TFV)\end{tabular}} & \multirow{2}{*}{TFV} & TabFact (\citeyear{TabFact}) & E, M & Wikipedia & Yes & 9K & 31K & 8K & 8K & 2440$\times$900 \\ 
 &  & InfoTabs (\citeyear{infotabs}) & W & Wikipedia & No & 1.9K & 18K & / & / & 792$\times$880 \\ 
\bottomrule
\end{tabular}
}
\caption{Statistics of our constructed train datasets, which are sampled from the original datasets. W, E, and M represent Web page, Excel, and Markdown tables, respectively. The symbols \# T and \# Q indicate the number of tables and questions, and Avg. means average.  }
\label{tab:dataset_statistics}
\vspace{-4mm}
\end{table*}

For training data generation, each expanded solution is randomly divided into two segments at position $j \sim \text{Uniform}\{1,\ldots,m-1\}$. The first segment, called \textbf{Hints}, includes the initial reasoning steps and is represented as $[s_1, \dots, s_j]$ The remaining steps $S_r=[s_{j+1}, \dots, s_m]$ are referred to as the \textbf{Completions}. These hints, when combined with the original question $Q$, constitute the input query $Q_r = [Q, s_1, \dots, s_j]$. By default, a long solution can generate three hint-completion pairs. The full dataset for this stage, $D_r$, is thus composed of tuples $(I, Q_r, S_r)$. For the reasoning task, the LVLM is expected to first perform step-by-step reasoning before generating the final answer. The full prompt is provided in Table \ref{tab:prompt-for-grpo} in Appendix \ref{sec:prompt-for-grpo}.

\begin{table*}[t] %
\centering
\scalebox{0.76}{
\begin{tabular}{lccccccccc} %
\toprule
\multirow{2}{*}{\textbf{Method}} & \multicolumn{1}{c}{\multirow{2}{*}{\textbf{Resolution}}} & \multicolumn{4}{c}{\textbf{Question Answering}} & \multicolumn{2}{c}{\textbf{Fact Verification}}  & \multicolumn{1}{c}{\multirow{2}{*}{\textbf{Avg. I.}}}  & \multicolumn{1}{c}{\multirow{2}{*}{\textbf{Avg. O.}}} \\
\cmidrule(lr){3-6} \cmidrule(lr){7-8}
 & \multicolumn{1}{c}{} & \multicolumn{1}{c}{\textbf{TabMWP}\textsubscript{\textit{I}}} & \multicolumn{1}{c}{\textbf{WTQ}\textsubscript{\textit{I}}} & \multicolumn{1}{c}{\textbf{HiTab}\textsubscript{\textit{I}}} & \multicolumn{1}{c}{\textbf{TAT-QA}\textsubscript{\textit{O}}} & \multicolumn{1}{c}{\textbf{TabFact}\textsubscript{\textit{I}}} & \multicolumn{1}{c}{\textbf{InfoTabs}\textsubscript{\textit{O}}} & \multicolumn{1}{c}{} & \multicolumn{1}{c}{}   \\  %
\hline
\multicolumn{10}{l}{{\cellcolor[rgb]{0.957,0.957,0.957}}\textit{Closed-Source LVLM}} \\

OpenAI-o4-mini & UNK & 86.70 & 78.20 & 44.80 & 56.16 & 84.70 & 78.30 & 73.60 & \textbf{67.23} \\
GPT-4o & UNK & 87.59 & 64.39 & 39.32 & 53.85 & 73.33 & \textbf{79.50} & 66.16 & 66.68 \\
Gemini-2.5-Pro & UNK & \textbf{89.90} & \textbf{80.34} & \textbf{46.44} & 56.29 & \textbf{85.02} & 77.15 & \textbf{75.43} & 66.72 \\
Claude-3.5-Sonnet & UNK & 83.30 & 71.80 & 41.31 & \textbf{59.33} & 60.08 & 70.30 & 64.12 & 64.82 \\
\hline
\multicolumn{10}{l}{{\cellcolor[rgb]{0.957,0.957,0.957}}\textit{Open-Source LVLM}} \\
Qwen2-VL-2B & Dyn. & 46.10 & 22.30 & 22.90 & 30.44 & 8.90 & 24.60 & 25.05 & 27.52  \\
DeepSeek-VL2 4.5B & Dyn. & 53.75 & 40.42 & 18.89 & 24.11 & 13.65 & 24.79 & 31.68 & 24.45  \\
mPLUG-Owl2 7B & 448 & 6.83 & 0.67 & 0.13 & 0.39 & 8.21 & 26.19 & 3.96 & 13.29 \\
Monkey 7B & 896 & 13.26 & 19.07 & 6.41 & 12.31 & 22.56 & 22.11 & 15.33 & 17.21 \\
Qwen2-VL-7B-Ins & Dyn & 49.51 & 19.73 & 5.33 & 20.85 & 40.00 & 46.56 & 28.64 & 33.71 \\
Qwen2-VL-7B & Dyn. & 63.80 & 46.50 & 33.10 & 46.50 & 7.40 & 32.80 & 37.70 & 39.65 \\
LLaVA-v1.5 7B & 336 & 6.05 & 1.24 & 2.03 & 2.97 & 18.9 & 28.31 & 7.06 & 15.64  \\
Table-LLaVA 7B  & 336 & 57.78 & 18.43 & 10.09 & 12.82$^{\dagger}$ & 59.85 & 65.26$^{\dagger}$ & 36.54 & 39.04   \\
Table-LLaVA 13B & 336 & 59.77 & 20.41 & 10.85 & 15.67$^{\dagger}$ & 65.00 & 66.91$^{\dagger}$ & 39.01 & 41.29 \\
Qwen2-VL-72B & Dyn. & 81.95 & 65.70 & 44.04 & 52.23 & 73.45 & 72.82 & 66.29 & 62.53 \\
QVQ-72B-Preview  & Dyn. & \textbf{86.20} & \textbf{68.20} & \textbf{45.70} & \textbf{55.48} & \textbf{77.68} & \textbf{74.64} & \textbf{69.45} & \textbf{65.06} \\
\hline
\multicolumn{10}{l}{{\cellcolor[rgb]{0.957,0.957,0.957}}\textit{Optimizated LVLM}} \\
Qwen2-VL-2B-SFT &  Dyn. & 70.00 & 31.00 & 31.90 & 19.69 & 51.70 & 37.40 & 46.15 & 28.55 \\
Qwen2-VL-2B-GRPO & Dyn. & 71.40 & \textbf{35.30} & 35.20 & \textbf{29.02} & 22.70 & 29.00 & 41.15 & 29.01  \\
Qwen2-VL-2B-\ours & Dyn. & \textbf{83.20} & 34.40  & \textbf{37.30} & 26.42 & \textbf{60.90} & \textbf{43.60} & \textbf{53.95} & \textbf{35.01}  \\
Qwen2-VL-7B-SFT & Dyn. & 90.30 & 46.80 & 48.50 & 37.82 & 73.20 & 57.60 & 64.70 & 47.71  \\
Qwen2-VL-7B-GRPO & Dyn. & 89.20 & \textbf{53.20} & 54.70 & \textbf{51.68} & 11.90 & 41.60 & 52.25 & 46.64 \\
Qwen2-VL-7B-\ours & Dyn. & \textbf{92.60} & 50.30 & \textbf{58.20} & 48.06 & \textbf{73.40} & \textbf{62.80} & \textbf{68.63} & \textbf{55.43} \\
\hline
\end{tabular}
}
\caption{Evaluation results on 4 held-in and 2 held-out multimodal tabular tasks. The subscripts $_I$ and $_O$ denote held-in and held-out, respectively, while $\dagger$ indicates the model has been trained on this dataset. ``Dyn.'' denotes dynamic resolution processing, where input images are adaptively resized to preserve aspect ratios. } 
\label{tab:main_result}
\vspace{-4mm}
\end{table*}

The reward function consists of two components: an accuracy reward $R_{\text{acc}}$ and a format reward $R_{\text{format}}$, combined as follows:
\begin{align}
    R_{\text{r}} = R_{\text{acc}} + R_{\text{format}}.
\end{align}
\textbf{Accuracy reward} is calculated by comparing the model-generated answer $MA$, extracted from within \texttt{<answer></answer>} tags, with the golden answer $GA$, using a binary reward scheme:
\begin{align}
    R_{\text{acc}} =
        \begin{cases}
        1, & \text{if  $MA = GA$} \\
        0, & \text{otherwise}
        \end{cases}
\end{align}
To ensure rigorous and consistent assessment, we adopt the widely used Math-Verify\footnote{\url{https://github.com/huggingface/Math-Verify}} library, which provides a standardized method for parsing and verifying mathematical expressions and numerical values when comparing $MA$ and $GA$.
\textbf{Format reward} incentivizes the model to organize its output correctly by placing the reasoning within \texttt{<think></think>} tags and the final answer within \texttt{<answer></answer>} tags. This is checked via regular expression matching ($REM$):
\begin{align}
    R_{\text{format}} =
        \begin{cases}
        1, & \text{if $REM(S_r) = True$} \\
        0, & \text{$$otherwise$$}
        \end{cases}
\end{align}

\paragraph{Unified GRPO-Based Training Objective.} For both PA-GRPO and HC-GRPO, we employ a unified policy optimization strategy, differing only in the reward definition. Following \citet{shao2024deepseekmath}, after computing the reward $R^i$ for each output $S^i$ in a rollout, the advantage is calculated as:
\begin{align}
    \hat{A}^i &= \frac{R^i - \text{mean}(\{R^1, \dots, R^G\})}{\text{std}(\{R^1, \dots, R^G\})},
    \label{eq:adv}
\end{align}
which normalizes reward relative across the group.

The overall loss function combines a clipped surrogate objective with a KL divergence penalty:
\begin{align}
    \mathcal{L}_{\text{GRPO}}(\theta) = \mathcal{L}_{\text{clip}}(\theta) - \beta \mathbb{D}_{\text{KL}}[\pi_\theta \| \pi_{\text{ref}}],
    \label{eq:loss-grpo}
\end{align}
where $\mathcal{L}_{\text{clip}}(\theta)$ adopts the proximal policy optimization mechanism:
\begin{align}
\mathcal{L}_{\text{clip}}(\theta) &= \frac{1}{G} \sum_{i=1}^{G} 
\min \Biggl(
    \frac{\pi_\theta(S^i | Q,I)}{\pi_{\theta_{\text{old}}}(S^i | Q,I)} \hat{A}^i, \nonumber\\
&\hspace{-4em}\quad\;\,
    \text{clip} \Bigl( 
        \frac{\pi_\theta(S^i | Q,I)}{\pi_{\theta_{\text{old}}}(S^i | Q,I)},\;
        1{-}\epsilon,\; 1{+}\epsilon 
    \Bigr) \hat{A}^i 
\Biggr),
\label{loss-clip}
\end{align}
constraining policy model updates to prevent destructive parameter changes. The KL divergence term $\mathbb{D}_{\text{KL}}$ regularizes the policy model $\pi_\theta$ to maintain proximity to the reference model $\pi_{\text{ref}}$, which is initialized as a frozen copy of the pretrained policy model. This dual mechanism balances reward maximization with behavioral consistency, mitigating catastrophic forgetting during GRPO training.

\section{Experiments}
\label{sec:exp}

\subsection{Baselines}

We compare the performance of \ours{} against the following baselines: 
(1) \textit{Open-source LVLMs}: Qwen2-VL \cite{Qwen2-VL}, Qwen2-VL-7B-Ins \cite{yang2025doestablesourcematter}, DeepSeek-VL2 \cite{deepseekvl2}, LLaVA v1.5 \cite{liu2023improvedllava}, Table-LLaVA \cite{zheng-etal-2024-table-llava}, mPLUG-Owl2 \cite{ye2023mplugowl2}, Monkey \cite{li2023monkey}, and QVQ-72B-Preview \cite{qvq-72b-preview}. (2) \textit{Closed-source LVLMs}: GPT-4o \cite{openai2024gpt4technicalreport}, Gemini 2.5 Pro \cite{team2023gemini} and Claude-3.5-Sonnet \cite{anthropic2024claude}. (3) Two model optimization methods: SFT and GRPO. 

\subsection{Datasets and Evaluation}

We conduct experiments on MMTab \cite{zheng-etal-2024-table-llava}, which is a recent large-scale dataset focused on multimodal table understanding tasks. We have chosen to exclude table-to-text tasks from our study, since they involve open-ended questions without fixed or definitive answers. The detailed statistics for our training data are presented in Table \ref{tab:dataset_statistics}, covering WTQ, HiTab, TabMWP, and TabFac. To assess the robustness of various optimization methods, we set aside similar tasks, such as TAT-QA, InfoTabs, as held-out. 
To evaluate performance, we employ the accuracy to evaluate overall reasoning performance and TEDs for perception evaluation.

Due to imbalanced dataset sizes, an equal number of entries from each are sampled. 
Our sampling process is carefully designed to create representative and manageable subsets for training and evaluation, particularly considering the computational constraints imposed by high-resolution images and long text sequences. Our procedure involves a three-step approach:

\noindent \textbf{Image Resolution Filter:} We exclude images with exceptionally large dimensions that could lead to out-of-memory errors. Specifically, any image with a total pixel count exceeding 1/8 of the Qwen2-VL model's maximum capacity (12,845,056 pixels) is removed.

\noindent \textbf{Output Length Filter:} For the PA-GRPO task, samples where the ground-truth structured texts (e.g., Markdown) exceed 2048 tokens are removed to prevent memory issues when generating very large tables.

\noindent \textbf{Random Sampling:} We proceed with the random sampling. First, we select a diverse set of table images from this filtered pool. Then, we sample questions corresponding to these chosen tables until our target of 8,000 entries is reached.

\subsection{Experimental Setup}
\label{sec:experimental_setup}
We strategically select two open-source LVLMs as our policy model: Qwen2-VL-2B and Qwen2-VL-7B \cite{Qwen2-VL}, since they have strong cognitive behaviors that enhance self-reflection on reasoning tasks \cite{gandhi2025cognitivebehaviorsenableselfimproving}. 
All experiments are conducted on 8 NVIDIA A100 80GB Tensor Core GPUs with DeepSpeed~\cite{rajbhandari2020zero,rasley2020deepspeed}, Zero stage 2, and HuggingFace Accelerate~\cite{accelerate}.
During the warm-up stage, we use AdamW optimizer \cite{loshchilov2017decoupled} with a 10\% warm-up ratio and 1000 steps. Following prior work~\cite{chen2025r1v}, learning rates are set $2\text{e}^{-5}$ and $5\text{e}^{-6}$ respectively for Qwen2-VL-2B and Qwen2-VL-7B. Given the large image resolution shown in Table \ref{tab:dataset_statistics}, we set batch sizes to 2 and 1.

For the PA-GRPO and HC-GRPO stage, we perform 4 rollouts per question ($G = 4$) and set the sampling temperature to 1 to encourage diverse reasoning trajectories. The maximum sequence length is set to $L = 1024$, ensuring that the model can generate complete reasoning paths. Both the policy model and reference model are initialized from the model after the warm-up, with the reference model frozen during training. The epoch and batch size are set to 2 and 1. Following~\cite{chen2025r1v}, the KL divergence coefficient $\beta$ in Eq.~\ref{eq:loss-grpo} is set to 0.04 by default, and the learning rate for the policy model is set to $1 \text{e}^{-6}$ for both Qwen2-VL-2B and Qwen2-VL-7B.

\subsection{Table Reasoning Performance}
Table \ref{tab:main_result} depicts the comprehensive comparison of \ours{} against baselines. By analyzing the experimental results, we have the following findings:

\paragraph{Open-source Model Hierarchy.} Open-source models exhibit a clear performance hierarchy aligned with model size. Smaller models (Qwen2-VL-2B: Avg. I.=25.05\%, Avg. O.=27.52\%) significantly underperform their larger counterparts (QVQ-72B-Preview: Avg. I.=69.45\%, Avg. O.=65.06\%) by 44.40\% and 37.54\%, respectively. Notably, the 72B parameter class achieves performance comparable to closed-source models (GPT-4o: Avg. I=66.16\%, Avg. O=66.68\%), demonstrating the scalability of open architectures.

\paragraph{Closed-source Model Superiority.} Closed-source models consistently outperform open-source models across most datasets. For example, Gemini-2.5-pro (Avg. I.=75.43\%, Avg. O.=66.72\%) outperforms the best open-source model by 5.98\% in Avg. I. and 1.66\% in Avg. O., suggesting stronger multimodal table reasoning abilities. 

\begin{table}[t!]
\centering
\resizebox{\linewidth}{!}{ 
\begin{tabular}{lccc}
\toprule
\textbf{Dataset} & \textbf{Qwen2-VL-2B} & \textbf{Table-R1-2B} & \textbf{Table-LLaVA 7B} \\
\midrule
WTQ\textsubscript{\textit{I}} & 0.41 & \textbf{0.73} & 0.56 \\
TabMWP\textsubscript{\textit{I}} & 0.48 & \textbf{0.81} & 0.80 \\
TabFact\textsubscript{\textit{I}} & 0.63 & \textbf{0.93} & 0.40 \\
HiTab\textsubscript{\textit{I}} & 0.24 & \textbf{0.54} & 0.32 \\
InfoTabs\textsubscript{\textit{O}} & 0.16 & 0.56 & \textbf{0.74} \\
TAT-QA\textsubscript{\textit{O}} & 0.47 & \textbf{0.70} & 0.57 \\
\bottomrule
\end{tabular}
}
\caption{Table structure recognition performance (TEDS score) on four held-in (subscript $I$) and two held-out (subscript $O$) datasets. The best-performing score on each dataset is highlighted in \textbf{bold}. Table-R1-2B represents Qwen2-VL-2B-Table-R1. }
\label{tab:teds_r1_performance}
\end{table}

\paragraph{Optimization Methods Comparison.} All optimized LVLMs demonstrate significant improvements over baseline models. Notably, \ours{} achieves superior overall performance on both held-in and held-out. Specifically, for Qwen2-VL-7B, \ours{} outperforms SFT by 3.93\% and GRPO by 16.38\% on held-in. On held-out, it surpasses SFT by 7.72\% and GRPO by 8.79\%. This performance notably exceeds that of Table-LLaVA 13B and can be comparable with GPT-4o on held-in.

\paragraph{GRPO Sensitivity to Initial Capability.} While GRPO generally outperforms SFT across most QA tasks, a notable performance gap in table fact verification is observed. This discrepancy arises from the initial capabilities of the policy model. When it is significantly low (Qwen2-VL-7B: Avg. I.=7.40\% on TabFact$_I$), the rewards derived from group responses tend to approach zero. This situation leads to low standard deviations in Eq. \ref{eq:adv}, which hinders the convergence of the reinforcement learning. 

\subsection{Table Perception Performance}
As Table \ref{tab:teds_r1_performance} shows, our Qwen2-VL-2B-Table-R1 demonstrates competitive or superior performance on most datasets compared to Table-LLaVA 7B. We note that Table-LLaVA's performance is higher on InfoTabs. This is expected, as Table-LLaVA was explicitly trained on the InfoTabs dataset, whereas for Table-R1, this was a held-out dataset. Our strong performance on held-out tasks like TAT-QA underscores the robustness of our approach.

\subsection{Ablation Studies}
 
\begin{table}[t] %
\scalebox{0.69}{
\begin{tabular}{l|cc|cc} 
\hline
\multirow{2}{*}{\textbf{Method}} & \multicolumn{2}{c|}{\textbf{Question Answering}} & \multicolumn{2}{c}{\textbf{Fact Verification}} \\ 
\cline{2-5}
 & \multicolumn{1}{c}{\textbf{Avg. QA\textsubscript{\textit{I}}}} & \multicolumn{1}{c|}{\textbf{TAT-QA\textsubscript{\textit{O}}}} & \multicolumn{1}{c}{\textbf{TabFact\textsubscript{\textit{I}}}} & \multicolumn{1}{c}{\textbf{InfoTabs\textsubscript{\textit{O}}}} \\ 
\hline
\multicolumn{5}{c}{{\cellcolor[rgb]{0.957,0.957,0.957}}\textit{Qwen2-VL-2B}} \\
\ours & 51.37 & 26.42 & 60.90 & 43.60 \\
\ \ w/o Warm-up & 38.87 & 23.16 & 20.50 & 26.10 \\
\ \ $\triangle$ & {\cellcolor[rgb]{1,0.4,0.4}}-12.50 & {\cellcolor[rgb]{1,0.592,0.592}}-3.26 & {\cellcolor[rgb]{1,0.2,0.2}}-40.40 & {\cellcolor[rgb]{1,0.4,0.4}}-17.50 \\
\ \ w/o PA-GRPO & 50.60 & 26.20 & 60.20 & 42.90 \\
\ \ $\triangle$ & {\cellcolor[rgb]{1,0.914,0.914}}-0.77 & {\cellcolor[rgb]{1,0.914,0.914}}-0.22 & {\cellcolor[rgb]{1,0.914,0.914}}-0.70 & {\cellcolor[rgb]{1,0.914,0.914}}-0.70 \\
\ \ w/o HC-GRPO & 36.27 & 20.08 & 45.80 & 41.90 \\
\ \ $\triangle$ & {\cellcolor[rgb]{1,0.4,0.4}}-19.70 & {\cellcolor[rgb]{1,0.592,0.592}}-6.34 & {\cellcolor[rgb]{1,0.4,0.4}}-15.10 & {\cellcolor[rgb]{1,0.592,0.592}}-1.70 \\
\hline
\end{tabular}
}
\caption{Effectiveness Across Different Stages. We report the average performance across various benchmarks, where Avg. $QA_I$ denotes the average accuracy of three QA datasets. $\triangle$ denotes the performance gap between \ours{} and its variants.} 
\label{tab:ablation_stage}
\vspace{-4mm}
\end{table}

\begin{table}[t!]
\centering
\begin{footnotesize}
\resizebox{1.0\linewidth}{!}{
\begin{tabular}{c|ccccc}
\toprule
& \multicolumn{5}{c}{Number of HC splits per solution} \\
\midrule
Dataset & \quad 1 \quad & \quad 2 \quad & \quad 3 \quad & \quad 4 \quad \\
\midrule
\ \ \ \quad TabMWP \ \ \ \ \quad & $81.5$ & $82.6$ & $83.2$ & $83.4$ \\
\bottomrule
\end{tabular}
}
\vspace{4mm}

\resizebox{1.0\linewidth}{!}{
\begin{tabular}{c|ccccc}
\toprule
& \multicolumn{5}{c}{Number of generations $G$ per question} \\
\midrule
Dataset & \quad 2 \quad & \quad 3 \quad & \quad 4 \quad & \quad 5 \quad & \quad 6 \quad \\
\midrule
\ \ \ \quad TabMWP \ \ \ \ \quad & $81.2$ & $82.5$ & $83.2$ & $83.4$ & $83.5$ \\
\bottomrule
\end{tabular}
}
\end{footnotesize}
\caption{Impact of the number of splits per solution and generations per question on TabMWP's performance. Experiments are conducted on Qwen2-VL-2B.}
\label{tab:effect_number_G&HC}
\end{table}

\begin{figure}[t]
    \centering
    \includegraphics[width=\linewidth]{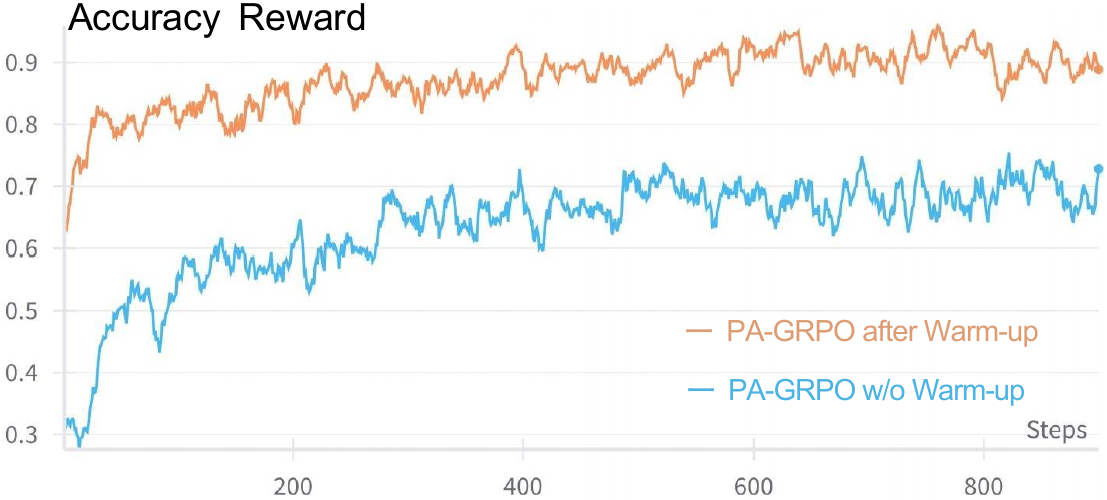}
    \caption{accuracy reward of PA-GRPO with and without warm-up.}
    \label{fig:table-r1-vs-grpo}
\end{figure}

\begin{table}[t]
\centering
\resizebox{\linewidth}{!}{
\begin{tabular}{lccc}
\toprule
\textbf{Dataset} & \textbf{Qwen2-VL-2B-Table-R1} & \textbf{w/o PA-GRPO} & \textbf{$\Delta$} \\
\midrule
WTQ\textsubscript{\textit{I}}        & 0.73 & 0.63 & -0.10 \\
TabMWP\textsubscript{\textit{I}}     & 0.81 & 0.53 & -0.28 \\
TabFact\textsubscript{\textit{I}}    & 0.93 & 0.85 & -0.08 \\
HiTab\textsubscript{\textit{I}}      & 0.54 & 0.41 & -0.13 \\
InfoTabs\textsubscript{\textit{O}}   & 0.56 & 0.53 & -0.03 \\
TAT-QA\textsubscript{\textit{O}}     & 0.70 & 0.62 & -0.08 \\
\bottomrule
\end{tabular}
}
\caption{Effectiveness of the PA-GRPO module on table perception performance of four held-in (subscript $I$) and two held-out (subscript $O$) datasets. 'w/o PA-GRPO' is the model variant without PA-GRPO, and $\Delta$ quantifies the resulting performance drop.}
\label{tab:effectiveness_pa_grpo}
\end{table}

\begin{table}[t!]
\centering
\begin{footnotesize}
\resizebox{1.0\linewidth}{!}{
\begin{tabular}{c|ccc}
 \toprule
 \textbf{Methods} & \quad \textbf{Warm-up(W)} \quad & \quad \textbf{W+PA-GRPO} \quad & \quad \textbf{W+SFT} \quad \\
 \midrule
\ \ \ \quad TEDS \ \ \ \ \quad  & $0.51$ & $0.80$  & $0.84$ \\
\ \ \ \quad Accuracy \ \ \ \ \quad  & $60.2$ & $58.7$ & $14.30$ \\
\bottomrule
\end{tabular}
}
\end{footnotesize}
\caption{Comparison of the warm-up, PA-GRPO, and SFT methods over TEDS and accuracy on TabMWP, based on training the Qwen2-VL-2B model.}
\label{tab:effect_ocr_grpo}
\end{table}

\begin{table}[t]
\centering
\resizebox{\linewidth}{!}{
\begin{tabular}{c|cccc}
\toprule
\textbf{Methods} & \textbf{W\textsubscript{qs}} & \textbf{W\textsubscript{qs}+GRPO} & \textbf{W\textsubscript{hc}} & \textbf{W\textsubscript{hc}+HC-GRPO} \\
\midrule
Accuracy & $60.20$ & $76.40$ & $63.60$ & \textbf{$83.00$} \\
\bottomrule
\end{tabular}
}
\caption{Effectiveness of the different methods on TabMWP using Qwen2-VL-2B, where $W_{qs}$ and $W_{hc}$ denote warm-up stage with question-solution pairs and hint-completion pairs.}
\label{tab:HC_GRPO}
\end{table}

\paragraph{Effects of the Number of $G$ and HC Splits.}
We conduct the parameter analysis on the number of generations $G$ in HC-GRPO with Qwen2-VL-2B over TabMWP$_I$, analyzing its impact on reasoning performance. Table~\ref{tab:effect_number_G&HC} demonstrates that a larger $G$ typically results in better performance, as the baseline reward is estimated as the average reward of all generated reasoning paths. A larger $G$ leads to low variance and a more stable estimation of the baseline reward, making the optimization process more stable. However, increasing $G$ also raises higher computational costs. Thus, $G=4$ is set as the default to balance performance and computational efficiency. Similarly, increasing the number of HC splits enhances performance by generating more training data, with a default value of 3.

\paragraph{Effectiveness of Warm-up.}
Table~\ref{tab:ablation_stage} reveals that the impact of excluding the warm-up phase causes the most severe performance drop (e.g., -40.40\% on TabFact$_I$), highlighting its necessity in mitigating GRPO's sensitivity to poor initial accuracy. Furthermore, integrating a warm-up stage consistently markedly improves accuracy rewards, as demonstrated by the steep initial increase in accuracy rewards in Figure \ref{fig:table-r1-vs-grpo}. This enhancement can be attributed to the rapid acquisition of fundamental reasoning capability during the warm-up stage.

\paragraph{Effectiveness of the PA-GRPO.}
Table~\ref{tab:ablation_stage} indicates that the absence of PA-GRPO yields only marginal performance changes. We conduct a comprehensive comparison of PA-GRPO and the standard SFT method on TabMWP, utilizing Qwen2-VL-2B as the foundational model. As shown in Table~\ref{tab:effect_ocr_grpo}, both TEDS and accuracy are used to assess recognition and reasoning abilities, respectively. Although SFT improves the model's visual recognition (TEDS 0.84), it severely impairs reasoning accuracy from 60.2\% to 14.30\%. 
In contrast, PA-GRPO significantly boosts recognition performance while maintaining strong reasoning capabilities, demonstrating its effectiveness as a more balanced optimization strategy. 
More details on the effectiveness of the PA-GRPO module on table perception performance are shown in Table \ref{tab:effectiveness_pa_grpo}. Removing PA-GRPO causes a substantial drop in accuracy across all datasets (e.g., a 0.28 drop on TabMWP). This result directly confirms that PA-GRPO is highly effective at its intended task: improving the model's ability to accurately recognize and represent table structures.

\paragraph{Effectiveness of the HC-GRPO.}
Our proposed HC-GRPO introduces a fine-grained, residual-step reward in contrast to conventional coarse solution-level rewards. To assess its effectiveness, we conduct a comparative study on TabMWP using Qwen2-VL-2B under four training configurations: 
(1) $\text{warm-up}_{qs}$ with question-solution pairs datasets; (2) $\text{warm-up}_{qs}$ + GRPO (solution-level reward); (3) $\text{warm-up}_{hc}$ using hint-completion pairs datasets; (4)  $\text{warm-up}_{hc}$ + HC-GRPO (residual-step reward).
As shown in Table~\ref{tab:HC_GRPO}, both $\text{warm-up}_{qs}$ and $\text{warm-up}_{hc}$ can improve performance. However, HC-GRPO can achieve higher accuracy. This reveals that residual-step rewards are more effective in enhancing LVLMs’ reasoning capabilities, as they offer finer-grained supervision by aligning rewards with the remaining reasoning steps, thereby enabling more precise credit assignment than coarse solution-level rewards.

\section{Conclusion}
We introduce \ours{}, a novel three-stage framework that significantly enhances multimodal table perception and reasoning by integrating warm-up initialization, continuous reward refinement through PA-GRPO, and fine-grained hint-based reasoning with HC-GRPO. Through extensive evaluation, \ours{} demonstrates superior performance and robustness compared to both SFT and GRPO methods. Additionally, it significantly outperforms existing open-source LVLM, even on par with the powerful GPT-4o on some benchmarks. Overall, our approach not only underscores the pivotal role of initial policy accuracy in reinforcement learning for reasoning tasks but also establishes a practical pathway for advancing RL-driven multimodal comprehension in real-world applications.

\section*{Limitations}
Despite its promising performance, our framework faces three key limitations that motivate future research. First, the current framework focuses primarily on generating definitive answers, leaving significant room for exploration in the area of table text generation. For example, tasks such as table summarization and table description generation are not fully addressed. Second, our evaluation relies on English‐only benchmarks with clear images, whereas real‐world table images often exhibit perspective distortions, uneven lighting, or handwriting, and multilingual contexts remain unaddressed. Third, the HC-GRPO stage relies on coarse binary rewards for correctness and formatting; richer signals such as step-level validity scores or continuous semantic similarity metrics could yield more nuanced training.

\section*{Acknowledgments}
We thank all anonymous reviewers for their valuable comments. The work was partially supported by the following: National Natural Science Foundation of China under No. 92370119, 62436009, 62276258 and 62376113, XJTLU Funding REF-22-01-002, Research Development Fund with No. RDF-22-01-020, and Suzhou Municipal Key Laboratory for Intelligent Virtual Engineering (SZS2022004). We would like to express our deepest gratitude to the Red Bird MPhil Program at the Hong Kong University of Science and Technology (Guangzhou) for providing us with generous support, resources, and funding.
 
\section*{Ethical Considerations}

We discuss the following ethical considerations related to Table-R1: (1) \textbf{Intellectual Property.} We adhere to the license when using existing datasets, such as Apache-2.0 for MMTab and MIT for TabMWP. (2) \textbf{Intended Use.} Table-R1 can be utilized to develop more persuasive multimodal table reasoning models. Researchers can also inherit our methodological design to develop their RL models in other scenarios. (3) \textbf{Controlling Potential Risks.} Since the training of Table-R1 only includes public datasets, which do not require extensive judgments about social risks, we believe Table-R1 does not introduce any additional risks. We manually verified some randomly sampled data from the experimental datasets to ensure the dataset did not contain risky issues.

\bibliography{custom}

\begin{thebibliography}{51}
\providecommand{\natexlab}[1]{#1}

\bibitem[{Anthropic(2024)}]{anthropic2024claude}
AI~Anthropic. 2024.
\newblock The claude 3 model family: Opus, sonnet, haiku.
\newblock \emph{Claude-3 Model Card}, 1:1.

\bibitem[{Chen et~al.(2025)Chen, Li, Zhao, Song, and Vinci}]{chen2025r1v}
Liang Chen, Lei Li, Haozhe Zhao, Yifan Song, and Vinci. 2025.
\newblock R1-v: Reinforcing super generalization ability in vision-language models with less than \$3.
\newblock \url{https://github.com/Deep-Agent/R1-V}.
\newblock Accessed: 2025-02-02.

\bibitem[{Chen et~al.(2020)Chen, Wang, Chen, Zhang, Wang, Li, Zhou, and Wang}]{TabFact}
Wenhu Chen, Hongmin Wang, Jianshu Chen, Yunkai Zhang, Hong Wang, Shiyang Li, Xiyou Zhou, and William~Yang Wang. 2020.
\newblock \href {https://arxiv.org/abs/1909.02164} {Tabfact: A large-scale dataset for table-based fact verification}.
\newblock \emph{Preprint}, arXiv:1909.02164.

\bibitem[{Chen et~al.(2021)Chen, Chen, Smiley, Shah, Borova, Langdon, Moussa, Beane, Huang, Routledge, and Wang}]{chen2021finqa}
Zhiyu Chen, Wenhu Chen, Charese Smiley, Sameena Shah, Iana Borova, Dylan Langdon, Reema Moussa, Matt Beane, Ting-Hao Huang, Bryan Routledge, and William~Yang Wang. 2021.
\newblock Finqa: A dataset of numerical reasoning over financial data.
\newblock \emph{Proceedings of EMNLP 2021}.

\bibitem[{Cheng et~al.(2024)Cheng, Li, Xu, Zhang, Zhou, and Liu}]{chengVisionLanguageModelsCan2024}
Kanzhi Cheng, Yantao Li, Fangzhi Xu, Jianbing Zhang, Hao Zhou, and Yang Liu. 2024.
\newblock \href {https://doi.org/10.48550/arXiv.2411.00855} {Vision-{{Language Models Can Self-Improve Reasoning}} via {{Reflection}}}.
\newblock \emph{Preprint}, arXiv:2411.00855.

\bibitem[{Cheng et~al.(2022)Cheng, Dong, Wang, Jia, Guo, Gao, Han, Lou, and Zhang}]{HiTab}
Zhoujun Cheng, Haoyu Dong, Zhiruo Wang, Ran Jia, Jiaqi Guo, Yan Gao, Shi Han, Jian-Guang Lou, and Dongmei Zhang. 2022.
\newblock \href {https://doi.org/10.18653/v1/2022.acl-long.78} {{H}i{T}ab: A hierarchical table dataset for question answering and natural language generation}.
\newblock In \emph{Proceedings of the 60th Annual Meeting of the Association for Computational Linguistics (Volume 1: Long Papers)}, pages 1094--1110, Dublin, Ireland. Association for Computational Linguistics.

\bibitem[{Chu et~al.(2025)Chu, Zhai, Yang, Tong, Xie, Schuurmans, Le, Levine, and Ma}]{chu2025sftmemorizesrlgeneralizes}
Tianzhe Chu, Yuexiang Zhai, Jihan Yang, Shengbang Tong, Saining Xie, Dale Schuurmans, Quoc~V. Le, Sergey Levine, and Yi~Ma. 2025.
\newblock \href {https://arxiv.org/abs/2501.17161} {Sft memorizes, rl generalizes: A comparative study of foundation model post-training}.
\newblock \emph{Preprint}, arXiv:2501.17161.

\bibitem[{Gandhi et~al.(2025)Gandhi, Chakravarthy, Singh, Lile, and Goodman}]{gandhi2025cognitivebehaviorsenableselfimproving}
Kanishk Gandhi, Ayush Chakravarthy, Anikait Singh, Nathan Lile, and Noah~D. Goodman. 2025.
\newblock \href {https://arxiv.org/abs/2503.01307} {Cognitive behaviors that enable self-improving reasoners, or, four habits of highly effective stars}.
\newblock \emph{Preprint}, arXiv:2503.01307.

\bibitem[{Gugger et~al.(2022)Gugger, Debut, Wolf, Schmid, Mueller, Mangrulkar, Sun, and Bossan}]{accelerate}
Sylvain Gugger, Lysandre Debut, Thomas Wolf, Philipp Schmid, Zachary Mueller, Sourab Mangrulkar, Marc Sun, and Benjamin Bossan. 2022.
\newblock Accelerate: Training and inference at scale made simple, efficient and adaptable.
\newblock \url{https://github.com/huggingface/accelerate}.

\bibitem[{Guo et~al.(2025)Guo, Yang, Zhang, Song, Zhang, Xu, Zhu, Ma, Wang, Bi et~al.}]{guo2025deepseekr1}
Daya Guo, Dejian Yang, Haowei Zhang, Junxiao Song, Ruoyu Zhang, Runxin Xu, Qihao Zhu, Shirong Ma, Peiyi Wang, Xiao Bi, et~al. 2025.
\newblock Deepseek-r1: Incentivizing reasoning capability in llms via reinforcement learning.
\newblock \emph{arXiv preprint arXiv:2501.12948}.

\bibitem[{Gupta et~al.(2020)Gupta, Mehta, Nokhiz, and Srikumar}]{infotabs}
Vivek Gupta, Maitrey Mehta, Pegah Nokhiz, and Vivek Srikumar. 2020.
\newblock \href {https://www.aclweb.org/anthology/2020.acl-main.210} {{INFOTABS}: Inference on tables as semi-structured data}.
\newblock In \emph{Proceedings of the 58th Annual Meeting of the Association for Computational Linguistics}, pages 2309--2324, Online. Association for Computational Linguistics.

\bibitem[{Kang et~al.(2025)Kang, Wang, Jin, Wang, Huang, and Wang}]{kangTemplateDrivenLLMParaphrasedFramework2025}
Xiaoqiang Kang, Zimu Wang, Xiaobo Jin, Wei Wang, Kaizhu Huang, and Qiufeng Wang. 2025.
\newblock \href {https://doi.org/10.1609/aaai.v39i23.34607} {Template-driven llm-paraphrased framework for tabular math word problem generation}.
\newblock \emph{Proceedings of the AAAI Conference on Artificial Intelligence}, 39(23):24303--24311.

\bibitem[{Katsis et~al.(2022)Katsis, Chemmengath, Kumar, Bharadwaj, Canim, Glass, Gliozzo, Pan, Sen, Sankaranarayanan, and Chakrabarti}]{katsis2022ait}
Yannis Katsis, Saneem Chemmengath, Vishwajeet Kumar, Samarth Bharadwaj, Mustafa Canim, Michael Glass, Alfio Gliozzo, Feifei Pan, Jaydeep Sen, Karthik Sankaranarayanan, and Soumen Chakrabarti. 2022.
\newblock \href {https://doi.org/10.18653/v1/2022.naacl-industry.34} {{AIT-QA}: {Q}uestion answering dataset over complex tables in the airline industry}.
\newblock In \emph{Proceedings of the 2022 Conference of the North American Chapter of the Association for Computational Linguistics: Human Language Technologies: Industry Track}, pages 305--314, Hybrid: Seattle, Washington + Online. Association for Computational Linguistics.

\bibitem[{Li et~al.(2024{\natexlab{a}})Li, Wang, Xu, Wang, Feng, Kong, and Liu}]{li-etal-2024-multimodal-arxiv}
Lei Li, Yuqi Wang, Runxin Xu, Peiyi Wang, Xiachong Feng, Lingpeng Kong, and Qi~Liu. 2024{\natexlab{a}}.
\newblock \href {https://doi.org/10.18653/v1/2024.acl-long.775} {Multimodal {A}r{X}iv: A dataset for improving scientific comprehension of large vision-language models}.
\newblock In \emph{Proceedings of the 62nd Annual Meeting of the Association for Computational Linguistics (Volume 1: Long Papers)}, pages 14369--14387, Bangkok, Thailand. Association for Computational Linguistics.

\bibitem[{Li et~al.(2024{\natexlab{b}})Li, Yang, Liu, Ma, Zhang, Yang, Sun, Liu, and Bai}]{li2023monkey}
Zhang Li, Biao Yang, Qiang Liu, Zhiyin Ma, Shuo Zhang, Jingxu Yang, Yabo Sun, Yuliang Liu, and Xiang Bai. 2024{\natexlab{b}}.
\newblock Monkey: Image resolution and text label are important things for large multi-modal models.
\newblock In \emph{proceedings of the IEEE/CVF conference on computer vision and pattern recognition}.

\bibitem[{Liu et~al.(2024)Liu, Li, Li, and Lee}]{liu2023improvedllava}
Haotian Liu, Chunyuan Li, Yuheng Li, and Yong~Jae Lee. 2024.
\newblock Improved baselines with visual instruction tuning.
\newblock In \emph{Proceedings of the IEEE/CVF Conference on Computer Vision and Pattern Recognition (CVPR)}, pages 26296--26306.

\bibitem[{Liu et~al.(2025{\natexlab{a}})Liu, Wang, and et~al.}]{liuUnderstandingR1ZeroLikeTraining2025}
Yicheng Liu, Bowen Wang, and et~al. 2025{\natexlab{a}}.
\newblock Understanding r1: On zero-like training and reward-conditioned policy optimization.
\newblock \emph{arXiv preprint arXiv:2503.11234}.

\bibitem[{Liu et~al.(2025{\natexlab{b}})Liu, Sun, Zang, Dong, Cao, Duan, Lin, and Wang}]{liuVisualRFTVisualReinforcement2025}
Ziyu Liu, Zeyi Sun, Yuhang Zang, Xiaoyi Dong, Yuhang Cao, Haodong Duan, Dahua Lin, and Jiaqi Wang. 2025{\natexlab{b}}.
\newblock \href {https://doi.org/10.48550/arXiv.2503.01785} {Visual-{{RFT}}: {{Visual Reinforcement Fine-Tuning}}}.
\newblock \emph{Preprint}, arXiv:2503.01785.

\bibitem[{Loshchilov and Hutter(2017)}]{loshchilov2017decoupled}
Ilya Loshchilov and Frank Hutter. 2017.
\newblock \href {https://arxiv.org/abs/1711.05101} {Decoupled weight decay regularization}.
\newblock \emph{arXiv preprint arXiv:1711.05101}.

\bibitem[{Lu et~al.(2023)Lu, Qiu, Chang, Wu, Zhu, Rajpurohit, Clark, and Kalyan}]{lu2023tabmwp}
Pan Lu, Liang Qiu, Kai-Wei Chang, Ying~Nian Wu, Song-Chun Zhu, Tanmay Rajpurohit, Peter Clark, and Ashwin Kalyan. 2023.
\newblock Dynamic prompt learning via policy gradient for semi-structured mathematical reasoning.
\newblock In \emph{International Conference on Learning Representations (ICLR)}.

\bibitem[{Luong et~al.(2024)Luong, Zhang, Jie, Sun, Jin, and Li}]{luongReFTReasoningReinforced2024}
Trung~Quoc Luong, Xinbo Zhang, Zhanming Jie, Peng Sun, Xiaoran Jin, and Hang Li. 2024.
\newblock \href {https://doi.org/10.48550/arXiv.2401.08967} {{{ReFT}}: {{Reasoning}} with {{Reinforced Fine-Tuning}}}.
\newblock \emph{Preprint}, arXiv:2401.08967.

\bibitem[{Mathur et~al.(2024)Mathur, Bafna, Kartik, Khandelwal, Shrivastava, Gupta, Bansal, and Roth}]{mathur-etal-2024-knowledge}
Suyash~Vardhan Mathur, Jainit~Sushil Bafna, Kunal Kartik, Harshita Khandelwal, Manish Shrivastava, Vivek Gupta, Mohit Bansal, and Dan Roth. 2024.
\newblock \href {https://doi.org/10.18653/v1/2024.findings-emnlp.822} {Knowledge-aware reasoning over multimodal semi-structured tables}.
\newblock In \emph{Findings of the Association for Computational Linguistics: EMNLP 2024}, pages 14054--14073, Miami, Florida, USA. Association for Computational Linguistics.

\bibitem[{OpenAI et~al.(2024)OpenAI, Achiam, Adler, Agarwal, Ahmad, Akkaya, Aleman, Almeida, Altenschmidt, Altman et~al.}]{openai2024gpt4technicalreport}
OpenAI, Josh Achiam, Steven Adler, Sandhini Agarwal, Lama Ahmad, Ilge Akkaya, Florencia~Leoni Aleman, Diogo Almeida, Janko Altenschmidt, Sam Altman, et~al. 2024.
\newblock \href {https://arxiv.org/abs/2303.08774} {Gpt-4 technical report}.
\newblock \emph{Preprint}, arXiv:2303.08774.

\bibitem[{Ouyang et~al.(2022)Ouyang, Wu, Jiang, Almeida, Wainwright, Mishkin, Zhang, Agarwal, Slama, Ray, Schulman, Hilton, Kelton, Miller, Simens, Askell, Welinder, Christiano, Leike, and Lowe}]{ouyang2022training}
Long Ouyang, Jeff Wu, Xu~Jiang, Diogo Almeida, Carroll~L. Wainwright, Pamela Mishkin, Chong Zhang, Sandhini Agarwal, Katarina Slama, Alex Ray, John Schulman, Jacob Hilton, Fraser Kelton, Luke~E. Miller, Maddie Simens, Amanda Askell, Peter Welinder, Paul~Francis Christiano, Jan Leike, and Ryan~J. Lowe. 2022.
\newblock Training language models to follow instructions with human feedback.
\newblock In \emph{NeurIPS}.

\bibitem[{Pasupat and Liang(2015)}]{WTQ}
Panupong Pasupat and Percy Liang. 2015.
\newblock \href {https://doi.org/10.3115/v1/P15-1142} {Compositional semantic parsing on semi-structured tables}.
\newblock In \emph{Proceedings of the 53rd Annual Meeting of the Association for Computational Linguistics and the 7th International Joint Conference on Natural Language Processing (Volume 1: Long Papers)}, pages 1470--1480, Beijing, China. Association for Computational Linguistics.

\bibitem[{Peng et~al.(2025)Peng, Chris, Wang, Wei, Pei, Qiu, Jian, Hao, Pan, Xie, Ge, Zhuang, Song, Liu, and Zhou}]{pengSkyworkR1VPioneering2025}
Yi~Peng, Chris, Xiaokun Wang, Yichen Wei, Jiangbo Pei, Weijie Qiu, Ai~Jian, Yunzhuo Hao, Jiachun Pan, Tianyidan Xie, Li~Ge, Rongxian Zhuang, Xuchen Song, Yang Liu, and Yahui Zhou. 2025.
\newblock \href {https://doi.org/10.48550/arXiv.2504.05599} {Skywork {{R1V}}: {{Pioneering Multimodal Reasoning}} with {{Chain-of-Thought}}}.
\newblock \emph{Preprint}, arXiv:2504.05599.

\bibitem[{Rafailov et~al.(2023)Rafailov, Sharma, Mitchell, Manning, Ermon, and Finn}]{rafailov2023dpo}
Rafael Rafailov, Archit Sharma, Eric Mitchell, Christopher~D Manning, Stefano Ermon, and Chelsea Finn. 2023.
\newblock \href {https://arxiv.org/abs/2305.18290} {Direct preference optimization: Your language model is secretly a reward model}.
\newblock In \emph{Thirty-seventh Conference on Neural Information Processing Systems}.

\bibitem[{Rajbhandari et~al.(2020)Rajbhandari, Rasley, Ruwase, and He}]{rajbhandari2020zero}
Samyam Rajbhandari, Jeff Rasley, Olatunji Ruwase, and Yuxiong He. 2020.
\newblock \href {https://ieeexplore.ieee.org/abstract/document/9355301/} {Zero: Memory optimizations toward training trillion parameter models}.
\newblock In \emph{SC20: International Conference for High Performance Computing, Networking, Storage and Analysis}.

\bibitem[{Rasley et~al.(2020)Rasley, Rajbhandari, Ruwase, and He}]{rasley2020deepspeed}
Jeff Rasley, Samyam Rajbhandari, Olatunji Ruwase, and Yuxiong He. 2020.
\newblock \href {https://dl.acm.org/doi/abs/10.1145/3394486.3406703} {Deepspeed: System optimizations enable training deep learning models with over 100 billion parameters}.
\newblock In \emph{Proceedings of SIGKDD}.

\bibitem[{Schulman et~al.(2017)Schulman, Wolski, Dhariwal, Radford, and Klimov}]{PPO}
John Schulman, Filip Wolski, Prafulla Dhariwal, Alec Radford, and Oleg Klimov. 2017.
\newblock Proximal policy optimization algorithms.
\newblock \emph{arXiv:1707.06347}.

\bibitem[{Shao et~al.(2024)Shao, Wang, Zhu, Xu, Song, Bi, Zhang, Zhang, Li, Wu et~al.}]{shao2024deepseekmath}
Zhihong Shao, Peiyi Wang, Qihao Zhu, Runxin Xu, Junxiao Song, Xiao Bi, Haowei Zhang, Mingchuan Zhang, YK~Li, Y~Wu, et~al. 2024.
\newblock Deepseekmath: Pushing the limits of mathematical reasoning in open language models.
\newblock \emph{arXiv preprint arXiv:2402.03300}.

\bibitem[{Team et~al.(2023)Team, Anil, Borgeaud, Alayrac, Yu, Soricut, Schalkwyk, Dai, Hauth, Millican et~al.}]{team2023gemini}
Gemini Team, Rohan Anil, Sebastian Borgeaud, Jean-Baptiste Alayrac, Jiahui Yu, Radu Soricut, Johan Schalkwyk, Andrew~M Dai, Anja Hauth, Katie Millican, et~al. 2023.
\newblock Gemini: a family of highly capable multimodal models.
\newblock \emph{arXiv preprint arXiv:2312.11805}.

\bibitem[{Team(2024)}]{qvq-72b-preview}
Qwen Team. 2024.
\newblock \href {https://qwenlm.github.io/blog/qvq-72b-preview/} {Qvq: To see the world with wisdom}.

\bibitem[{Van~Breugel and Van Der~Schaar(2024)}]{pmlr-v235-van-breugel24a}
Boris Van~Breugel and Mihaela Van Der~Schaar. 2024.
\newblock \href {https://proceedings.mlr.press/v235/van-breugel24a.html} {Position: Why tabular foundation models should be a research priority}.
\newblock In \emph{Proceedings of the 41st International Conference on Machine Learning}, volume 235 of \emph{Proceedings of Machine Learning Research}, pages 48976--48993. PMLR.

\bibitem[{Wang et~al.(2024{\natexlab{a}})Wang, Li, Shao, Xu, Dai, Li, Chen, Wu, and Sui}]{wangMathShepherdVerifyReinforce2024}
Peiyi Wang, Lei Li, Zhihong Shao, R.~X. Xu, Damai Dai, Yifei Li, Deli Chen, Y.~Wu, and Zhifang Sui. 2024{\natexlab{a}}.
\newblock \href {https://arxiv.org/abs/2312.08935} {Math-{{Shepherd}}: {{Verify}} and {{Reinforce LLMs Step-by-step}} without {{Human Annotations}}}.
\newblock \emph{Preprint}, arXiv:2312.08935.

\bibitem[{Wang et~al.(2024{\natexlab{b}})Wang, Bai, Tan, Wang, Fan, Bai, Chen, Liu, Wang, Ge, Fan, Dang, Du, Ren, Men, Liu, Zhou, Zhou, and Lin}]{Qwen2-VL}
Peng Wang, Shuai Bai, Sinan Tan, Shijie Wang, Zhihao Fan, Jinze Bai, Keqin Chen, Xuejing Liu, Jialin Wang, Wenbin Ge, Yang Fan, Kai Dang, Mengfei Du, Xuancheng Ren, Rui Men, Dayiheng Liu, Chang Zhou, Jingren Zhou, and Junyang Lin. 2024{\natexlab{b}}.
\newblock Qwen2-vl: Enhancing vision-language model's perception of the world at any resolution.
\newblock \emph{arXiv preprint arXiv:2409.12191}.

\bibitem[{Wang et~al.(2025)Wang, Zhang, Zhang, Hu, Li, Zhang, Li, Wu, Wang, and Hovy}]{wangReinforcementLearningEnhanced2025}
Shuhe Wang, Shengyu Zhang, Jie Zhang, Runyi Hu, Xiaoya Li, Tianwei Zhang, Jiwei Li, Fei Wu, Guoyin Wang, and Eduard Hovy. 2025.
\newblock \href {https://doi.org/10.48550/arXiv.2412.10400} {Reinforcement {{Learning Enhanced LLMs}}: {{A Survey}}}.
\newblock \emph{Preprint}, arXiv:2412.10400.

\bibitem[{Wu et~al.(2024)Wu, Chen, Pan, Liu, Liu, Dai, Gao, Ma, Wu, Wang et~al.}]{deepseekvl2}
Zhiyu Wu, Xiaokang Chen, Zizheng Pan, Xingchao Liu, Wen Liu, Damai Dai, Huazuo Gao, Yiyang Ma, Chengyue Wu, Bingxuan Wang, et~al. 2024.
\newblock \href {https://arxiv.org/abs/2412.10302} {Deepseek-vl2: Mixture-of-experts vision-language models for advanced multimodal understanding}.
\newblock \emph{Preprint}, arXiv:2412.10302.

\bibitem[{Yang et~al.(2025)Yang, Zhang, Liu, Freitas, and Lin}]{yang2025doestablesourcematter}
Bohao Yang, Yingji Zhang, Dong Liu, André Freitas, and Chenghua Lin. 2025.
\newblock \href {https://arxiv.org/abs/2501.13042} {Does table source matter? benchmarking and improving multimodal scientific table understanding and reasoning}.
\newblock \emph{Preprint}, arXiv:2501.13042.

\bibitem[{Yang et~al.(2022)Yang, Gupta, Upadhyay, He, Goel, and Paul}]{yang-etal-2022-tableformer}
Jingfeng Yang, Aditya Gupta, Shyam Upadhyay, Luheng He, Rahul Goel, and Shachi Paul. 2022.
\newblock {TableFormer: Robust Transformer Modeling for Table-Text Encoding}.
\newblock In \emph{Proc. of ACL}.

\bibitem[{Ye et~al.(2023)Ye, Xu, Ye, Yan, Hu, Liu, Qian, Zhang, Huang, and Zhou}]{ye2023mplugowl2}
Qinghao Ye, Haiyang Xu, Jiabo Ye, Ming Yan, Anwen Hu, Haowei Liu, Qi~Qian, Ji~Zhang, Fei Huang, and Jingren Zhou. 2023.
\newblock \href {https://arxiv.org/abs/2311.04257} {mplug-owl2: Revolutionizing multi-modal large language model with modality collaboration}.
\newblock \emph{Preprint}, arXiv:2311.04257.

\bibitem[{Yu et~al.(2025)Yu, Zhang, Zhu, Yuan, Zuo, Yue, Dai, Fan, Liu, Liu et~al.}]{yuDAPOOpenSourceLLM2025}
Qiying Yu, Zheng Zhang, Ruofei Zhu, Yufeng Yuan, Xiaochen Zuo, Yu~Yue, Weinan Dai, Tiantian Fan, Gaohong Liu, Lingjun Liu, et~al. 2025.
\newblock \href {https://neurips.cc/virtual/2025/loc/san-diego/poster/120129} {Dapo: An open-source llm reinforcement learning system at scale}.
\newblock In \emph{Proceedings of the Thirty-Ninth Conference on Neural Information Processing Systems (NeurIPS 2025), Poster Session}, page Poster 120129.
\newblock Accepted and presented at NeurIPS 2025, San Diego, USA.

\bibitem[{Yuan et~al.(2025)Yuan, Pang, Cho, Li, Sukhbaatar, Xu, and Weston}]{yuanSelfRewardingLanguageModels2025}
Weizhe Yuan, Richard~Yuanzhe Pang, Kyunghyun Cho, Xian Li, Sainbayar Sukhbaatar, Jing Xu, and Jason Weston. 2025.
\newblock \href {https://doi.org/10.48550/arXiv.2401.10020} {Self-{{Rewarding Language Models}}}.
\newblock \emph{Preprint}, arXiv:2401.10020.

\bibitem[{Zhang et~al.(2024)Zhang, Zhoubian, Hu, Yue, Dong, and Tang}]{zhangReSTMCTSLLMSelfTraining2024}
Dan Zhang, Sining Zhoubian, Ziniu Hu, Yisong Yue, Yuxiao Dong, and Jie Tang. 2024.
\newblock \href {https://arxiv.org/abs/2406.03816} {{{ReST-MCTS}}*: {{LLM Self-Training}} via {{Process Reward Guided Tree Search}}}.
\newblock \emph{Preprint}, arXiv:2406.03816.

\bibitem[{Zhang et~al.(2025)Zhang, Huang, Yao, Liu, Zhang, Lu, and Tao}]{zhangR1VLLearningReason2025}
Jingyi Zhang, Jiaxing Huang, Huanjin Yao, Shunyu Liu, Xikun Zhang, Shijian Lu, and Dacheng Tao. 2025.
\newblock \href {https://doi.org/10.48550/arXiv.2503.12937} {R1-{{VL}}: {{Learning}} to {{Reason}} with {{Multimodal Large Language Models}} via {{Step-wise Group Relative Policy Optimization}}}.
\newblock \emph{Preprint}, arXiv:2503.12937.

\bibitem[{Zhao et~al.(2024)Zhao, Feng, Liu, Tang, Wu, Liao, Wei, Ye, Liu, Zhou, Li, and Huang}]{zhao2024tabpedia}
Weichao Zhao, Hao Feng, Qi~Liu, Jingqun Tang, Binghong Wu, Lei Liao, Shu Wei, Yongjie Ye, Hao Liu, Wengang Zhou, Houqiang Li, and Can Huang. 2024.
\newblock Tabpedia: Towards comprehensive visual table understanding with concept synergy.
\newblock In \emph{Advances in Neural Information Processing Systems}.

\bibitem[{Zheng et~al.(2024)Zheng, Feng, Si, She, Lin, Jiang, and Wang}]{zheng-etal-2024-table-llava}
Mingyu Zheng, Xinwei Feng, Qingyi Si, Qiaoqiao She, Zheng Lin, Wenbin Jiang, and Weiping Wang. 2024.
\newblock \href {https://doi.org/10.18653/v1/2024.acl-long.493} {Multimodal table understanding}.
\newblock In \emph{Proceedings of the 62nd Annual Meeting of the Association for Computational Linguistics (Volume 1: Long Papers)}, pages 9102--9124, Bangkok, Thailand. Association for Computational Linguistics.

\bibitem[{Zheng et~al.(2021)Zheng, Burdick, Popa, Zhong, and Wang}]{zheng2020FinTabNet}
Xinyi Zheng, Doug Burdick, Lucian Popa, Peter Zhong, and Nancy Xin~Ru Wang. 2021.
\newblock Global table extractor (gte): A framework for joint table identification and cell structure recognition using visual context.
\newblock \emph{Winter Conference for Applications in Computer Vision (WACV)}.

\bibitem[{Zhong et~al.(2020)Zhong, ShafieiBavani, and Jimeno~Yepes}]{Xu2020PubTabNet}
Xu~Zhong, Elaheh ShafieiBavani, and Antonio Jimeno~Yepes. 2020.
\newblock Image-based table recognition: Data, model, and evaluation.
\newblock In \emph{Computer Vision -- ECCV 2020}, pages 564--580, Cham. Springer International Publishing.

\bibitem[{Zhong et~al.(2019)Zhong, ShafieiBavani, and Yepes}]{zhong2019image}
Xu~Zhong, Elaheh ShafieiBavani, and Antonio~Jimeno Yepes. 2019.
\newblock Image-based table recognition: data, model, and evaluation.
\newblock \emph{arXiv preprint arXiv:1911.10683}.

\bibitem[{Zhu et~al.(2021)Zhu, Lei, Huang, Wang, Zhang, Lv, Feng, and Chua}]{zhu2021tatqa}
Fengbin Zhu, Wenqiang Lei, Youcheng Huang, Chao Wang, Shuo Zhang, Jiancheng Lv, Fuli Feng, and Tat-Seng Chua. 2021.
\newblock \href {https://doi.org/10.18653/v1/2021.acl-long.254} {{TAT}-{QA}: A question answering benchmark on a hybrid of tabular and textual content in finance}.
\newblock In \emph{Proceedings of the 59th Annual Meeting of the Association for Computational Linguistics and the 11th International Joint Conference on Natural Language Processing (Volume 1: Long Papers)}, pages 3277--3287, Online. Association for Computational Linguistics.

\end{thebibliography}

\appendix

\newpage

\section{Pseudocode of \ours{}}
\label{apd:Pseudocode}
The overall training algorithm of \ours{} is presented in Algorithm \ref{alg:table_r1_Pseudocode}.

\begin{algorithm}[h]
\caption{Pseudocode of our \ours{}}
\label{alg:table_r1_Pseudocode}
\begin{algorithmic}
    \small 
    \STATE {\bfseries Input:} Policy model $\pi_{\theta}$ initialized by a pre-trained LVLM; a vision-text dataset $D_{p}$ and $D_{r}$.
    \STATE {\bfseries Output:} Trained policy model $\pi_{\theta}$
    \STATE \textit{Policy warm-up:}
    \FOR{\textit{iter = 1 to $N$}}
    \STATE Sample $\{I, Q, S\} \in D_{s} \cup  D_p$ 
    \STATE Optimize policy model $\pi_{\theta}$ by SFT
    \ENDFOR
    \STATE \textit{Perception Stage:}
    \FOR{\textit{iter = 1 to $N$}}
    \STATE Sample $\{I, Q_p, S_p\} \in D_{p}$
    \STATE Generate a group of perception paths $\{\mathbf{c}^i\}^M_{i=1} \sim \pi_{\theta}$
    \STATE Obtain Tree-Edit-Distance-based Similarity (TEDS) as rewards $\{r^i\}_{i=1}^M$ 
    \STATE Obtain relative advantages $\{\hat{A}^i\}_{i=1}^M$ by Eq.~\ref{eq:adv} 
    \STATE Optimize policy model $\pi_{\theta}$ by Eq \ref{eq:loss-grpo} 
    \ENDFOR
    \STATE \textit{ Reasoning Stage:}
    \FOR{\textit{iter = 1 to $N$}}
    \STATE Sample $\{I, Q_r, S_r\} \in D_{r}$
    \STATE Generate a group of reasoning paths $\{\mathbf{c}^i\}^M_{i=1} \sim \pi_{\theta}$
    \STATE Obtain accuracy rewards and format rewards $\{r^i\}_{i=1}^M$ 
    \STATE Obtain relative advantages $\{\hat{A}^i\}_{i=1}^M$ by Eq.~\ref{eq:adv} 
    \STATE Optimize policy model $\pi_{\theta}$ by Eq \ref{eq:loss-grpo} 
    \ENDFOR
  \RETURN policy model $\pi_{\theta}$
\end{algorithmic}
\end{algorithm}

\section{More Details about \ours{}}
\subsection{Prompts for GRPO Training}
\label{sec:prompt-for-grpo}

To ensure correct output formatting during the training of PA-GRPO and HC-GRPO, we adopt the prompts presented in Table \ref{tab:prompt-for-grpo}.

\begin{table}[h]
\caption{Prompts used in PA-GRPO and HC-GRPO. }
\vspace{-2mm}
\label{tab:prompt-for-grpo}
\begin{center}
\setlength{\tabcolsep}{4pt}
\scalebox{0.8}
{
\begin{tabular}{p{9cm}}  %
\toprule
\textbf{Perception Prompt:} A conversation between User and Assistant. The user asks a question, and the Assistant solves it. This task is a simple perception task, and the Assistant directly provides the answer within the $<$answer$>$ $<$/answer$>$ tags. For example: $<$answer$>$ answer here $<$/answer$>$
\\
\midrule
\textbf{Hint-Completion Prompt:} A conversation between User and Assistant. The user asks a question, and the Assistant solves it. The assistant first thinks about the reasoning process in the mind and then provides the user with the answer. The reasoning process and answer are enclosed within $<$think$>$ $<$/think$>$ and $<$answer$>$ $<$/answer$>$ tags, respectively, i.e., $<$think$>$ reasoning process here $<$/think$>$$<$answer$>$ answer here without unit $<$/answer$>$
\\
\bottomrule
\end{tabular}
}
\vspace{-4mm}
\end{center}
\end{table}

\subsection{Instruction variants for PA-GRPO Task}
\label{sec:Instruction-variants-for-PA-GRPO}
In the stage of PA-GRPO in Section \ref{subsec:table_r1}, we generate various instructions to transform the image into structured content. All possible templates are listed in Figure \ref{fig:question-perception-tr}. During the construction of the $D_p$, we randomly select one from them. 

\begin{figure}[htbp]
    \begin{tcolorbox}[
    left=2mm,right=1mm,top=0mm, bottom=0mm,colback=white,colframe=black]
    \begin{lstlisting}[style=plain]
"Please read the table in this image and return a markdown-style reconstructed table in text.",
"Take a look at the table in this image and provide me with the markdown representation of the table in text format.",
"Read the shown table in this image and give me the reconstructed table in the markdown text format.",
"Watch the table in this image and convert it into a Markdown table in the text form.",
"Given a table image, can you convert the table into a Markdown table in text form?",
"Reconstruct the table in this picture as a markdown-style table in text.",
"Please review this table image and return a text representation of the table in the markdown format.",
"Examine the table in the shown picture and generate a markdown text representation of the table.",
"Watch this table and show a markdown-style reconstructed table in text.",
"This picture illustrates a table. Please represent this table with the markdown format in text.",
"Recognize the table in the presented picture and represent it in the markdown format.",
"Recognize the table in this picture and return a markdown-style reconstructed table in text.",
"Can you interpret the table in this image and return it as a markdown table in text?"
"Look at the table in this image and reconstruct it as a markdown table in text format."
"Identify the table in this image and provide its markdown text representation."
"Please examine the table in this image and return it as a markdown table in text format."
"Can you read the table in this image and give me the markdown table in text?"
"Please look at the table in this image and provide the markdown table in text format."
    \end{lstlisting}
    \end{tcolorbox}
    \caption{Instruction variants for constructing the table recognition task.}
    \label{fig:question-perception-tr}
\end{figure}

\subsection{Prompts for Long-COT Data Generation}
\label{sec:prompt-for-data-construction}
In the warm-up stage in Section \ref{subsec:table_r1}, some original solutions are too short to be divided into two parts. Therefore, we use the prompt, shown in Figure \ref{fig:prompt-data-rewriting}, to expand short-COT into long-COT formats.

\begin{figure}[h!]
    \begin{tcolorbox}[
    left=2mm,right=1mm,top=0mm, bottom=0mm,colback=white,colframe=black]
    \begin{lstlisting}[style=plain]
Instruction
Your role is to serve as a step-by-step solution provider for mathematical exercises. You will generate a detailed explanation of the solution to a given mathematical question, using the provided table or data. Your explanations should be clear, logical, and adhere to the following guidelines:
1. **Provide a Detailed Step-by-Step Solution**: Generate a comprehensive, step-by-step guide that explains each part of the solution process, including the reasoning behind each step.
2. **Maintain the Integrity of the Answer**: Ensure that the final answer provided in the explanation aligns with the one given in the input without altering its fundamental correctness.
Here is an example of how the input and output should be structured:
<an Example of Demonstration>

    \end{lstlisting}
    \end{tcolorbox}
    \caption{Prompt for constructing long-COT solution.}
    \label{fig:prompt-data-rewriting}
\end{figure}

\subsection{Qualitative Breakdown of Failure Cases}
Our error analysis identified five main categories of failures. In descending order of frequency, they are as follows:

\begin{itemize}
    \item \textbf{Basic Arithmetic Calculation Errors:} The model frequently makes fundamental mistakes in basic mathematical operations (e.g., addition, subtraction).
    \item \textbf{Failure to Parse Complex Data Formats:} The model struggles to correctly interpret complex tables, such as those containing merged cells, or tables that are exceptionally long or wide.
    \item \textbf{Misunderstanding of Boundary Conditions:} This involves the poor interpretation of qualifying phrases (e.g., "at least," "more than"), leading to incorrect filtering of data.
    \item \textbf{Core Concept Confusion:} In these cases, the model misunderstands a core mathematical concept required by the question, such as 'absolute value'.
    \item \textbf{Omission of Key Information:} The least common error, this happens when the model fails to process the entire user prompt, overlooking crucial sentences or data points.
\end{itemize}

\end{document}